\documentclass[11pt]{article}
\usepackage{coling2020}
\usepackage{times}
\usepackage{url}
\usepackage{latexsym}
\usepackage{graphicx}
\usepackage{array, caption}
\usepackage{makecell}
\usepackage[utf8]{inputenc}
\usepackage[T1]{fontenc}
\usepackage{array}
\usepackage{booktabs}
\usepackage{url}
\usepackage{wasysym}
\usepackage{pgfplots}

\usepackage[T1]{fontenc}
\usepackage[font=small,labelfont=bf,tableposition=top]{caption}

\usepackage{todonotes}
\usepackage[T1]{fontenc}
\usepackage[font=small,labelfont=bf,tableposition=top]{caption}

\DeclareCaptionLabelFormat{andtable}{#1~#2  \&  \tablename~\thetable}

\definecolor{azure(colorwheel)}{rgb}{0.0, 0.5, 1.0}
\definecolor{ao(english)}{rgb}{0.0, 0.5, 0.0}

\colingfinalcopy % Uncomment this line for the final submission

\title{C1 at SemEval-2020 Task 9: SentiMix: Sentiment Analysis for Code-Mixed Social Media Text using Feature Engineering}

\author{Laksh Advani \and Clement Lu \and Suraj Maharjan \\
 Capital One\\
 1600 Capital One Drive, McLean, VA\\
 {\tt firstname.lastname@capitalone.com} }

\date{}

\begin{document}
\maketitle
\begin{abstract}
In today's interconnected and multilingual world, code-mixing of languages on social media is a common occurrence. While many Natural Language Processing (NLP) tasks like sentiment analysis are mature and well designed for monolingual text, techniques to apply these tasks to code-mixed text still warrant exploration. This paper describes our feature engineering approach to sentiment analysis in code-mixed social media text for SemEval-2020 Task 9: SentiMix. We tackle this problem by leveraging a set of hand-engineered lexical, sentiment, and metadata features to design a classifier that can disambiguate between ``positive'', ``negative'' and ``neutral'' sentiment. With this model we are able to obtain a weighted F1 score of $0.65$ for the ``Hinglish'' task and $0.63$ for the ``Spanglish'' tasks.\let\thefootnote\relax\footnotetext{*Our codalab username is \textit{lakshadvani}.}
%~\cite{advani2020sentimix} 
\end{abstract}

\section{Introduction}
\blfootnote{
    %
    % for review submission
    %
    %
    % % final paper: en-uk version 
    %
    % \hspace{-0.65cm}  % space normally used by the marker
    % This work is licensed under a Creative Commons 
    % Attribution 4.0 International Licence.
    % Licence details:
    % \url{http://creativecommons.org/licenses/by/4.0/}.
    % 
    % % final paper: en-us version 
    %
     \hspace{-0.65cm}  % space normally used by the marker
     This work is licensed under a Creative Commons 
     Attribution 4.0 International License.
     License details:
     \url{http://creativecommons.org/licenses/by/4.0/}.
}

Social media has grown exponentially in the last decade and has given rise to multilingual online communication and the consumption of media. Code Mixing is the phenomenon of embedding linguistic units such as phrases, words, or morphemes of one language into an utterance of another. Code mixing is prevalent in online discourse, where bilingual speakers mix English with their native language. For instance, bilingual speakers of Hindi and Spanish languages, which are the 3rd and 4th most spoken languages in the world respectively, frequently mix English with Hindi and Spanish in spoken language as well as online social media. The growing presence of ``Hinglish” and ``Spanglish” in social media is evidence that this is a growing area of research as traditional NLP tasks like sentiment analysis heavily rely on monolingual resources.

While there are many cutting edge techniques to apply common NLP tasks to monolingual text, the task of sentiment analysis, in particular, has not been explored for multilingual code-mixed texts. Conventional NLP tools make use of monolingual resources to operate on code-mixed text, which limits them to properly handle issues like word-level code-mixing. Additionally, code-mixed texts typically exist on social media which has a conjunction of other features like incorrect spelling, use of slang, and abbreviations to name a few.

While code-mixing is an active area of research, correctly annotated data is still scarce. It is particularly difficult to mine a small subset of tweets that may or may not be code-mixed and then apply a sentiment score to them. The organizers have solved this by releasing around 20,000 tweets \cite{patwa2020sentimix} with unique sentiment scores and word-level tagging of the language. For evaluation and ranking, the organizers used a weighted F1 score across ``positive'', ``negative'' and ``neutral'' classes.

While deep learning techniques are state of the art for this task, the contribution of this paper is to demonstrate how content-rich features, when applied to a classifier, can give us strong and comparable results. Some of the features we explore are lexical features like $n$-grams, metadata features like word frequency, and sentiment-based features like profanity count. 

\section{Related Work}
Code-Mixed Sentiment Analysis is an active field of research. For monolingual sentiment analysis, Recurrent Neural Networks (RNN) and other more complex deep learning models have been successful. \newcite{socher-etal-2013-recursive} gave us a significant breakthrough by using compositional vector representations.
With regards to Hindi-English Sentiment Analysis, the shared task SAIL-2015~\cite{10.1007/978-3-319-26832-3_61} reported accuracies of different systems submitted on sentiment analysis for tweets in three major Indian languages: Hindi, Bengali, and Tamil. Recently \newcite{joshi-etal-2016-towards} released a dataset sourced from Facebook and proposed a deep learning approach to Hindi-English Sentiment extraction. A variety of traditional learning algorithms like Naive Bayes, Support Vector Machines and Decision Trees were applied to this dataset. The winning algorithm from the SAIL-2015 shared task was a Naive Bayes based system.
More recent work by \newcite{hashimoto-etal-2017-joint} involves a
hierarchical multitask neural architecture with the lower layers performing syntactic tasks, and the higher layers performing the more involved semantic tasks while using the lower layer predictions.
\section{Dataset}

\begin{table}[!htbp]
 \centering
 \begin{tabular}{|l|l|}
 \hline
 Word & Word Tag \\
 \hline
 @ & O\\
 nsfw & Hin\\
 \textunderscore & O\\
 hs & Hin\\
 Maybe & Hin\\
 I'll & Eng\\
 faing & Eng\\
 and & Eng\\
 I & Eng\\
 won't & Eng\\
 feel & Eng\\
 a & Eng\\
 thing & Eng\\
 . & O\\
 
 \hline
 \end{tabular}
 \begin{tabular}{|l|l|}
 \hline
 Word & Word Tag \\
 \hline
 Nobody & Eng\\
 can & Eng\\
 make & Eng\\
 gabby & NE\\
 laugh & Eng\\
 like & Eng\\
 I & Eng\\
 do & Eng\\
 \smiley & O\\
 asta & Es\\
 parece & Es\\
 mongolita & Es\\
 lmao & Eng\\
 & \\
 \hline
 \end{tabular}
 \caption{Sample Annotation Examples for Hinglish and Spanglish Dataset}
 \label{table:eg}
 \end{table}

%For each language the organizers have released a total of 20,000 manually labeled tweets with 14,000 being in the training dataset, 3,000 being part of the validation dataset and 3,000 being part of a holdout test dataset. 

For the Hinglish dataset, the organizers have released a total of 20,000 manually labeled tweets with 14,000 being in the training dataset, 3,000 being part of the validation dataset and 3,000 being part of a holdout test dataset. Additionally, for the Spanglish dataset, they released 12,002 tweets for the training dataset, 2,998 tweets for the validation dataset and 3,788 tweets for the test dataset.

To ensure the accuracy of the dataset labels, the organizers employ a semi-automatic annotation methodology. They initially use word-level identifiers and then baseline sentiment analysis to obtain the initial labels. Subsequently, manual evaluation is done by at least two annotators, tweets with low confidence are discarded. While this seems to be a robust system in some cases the word tags are incorrect with words like 'Maybe' being tagged as a Hindi word, when we can safely assume that it is from the English language. Additionally, the dataset contains URLs, hashtags, usernames and emoticons. The average length for tweets in the ``Hinglish'' dataset is 26 tokens and 15 tokens for the ``Spanglish'' dataset.

Table 1 shows us two examples from the ``Hinglish'' and ``Spanglish'' datasets made available for this task. For both subtasks the organizers provided labeled trial, train and validation datasets for developing candidate models.

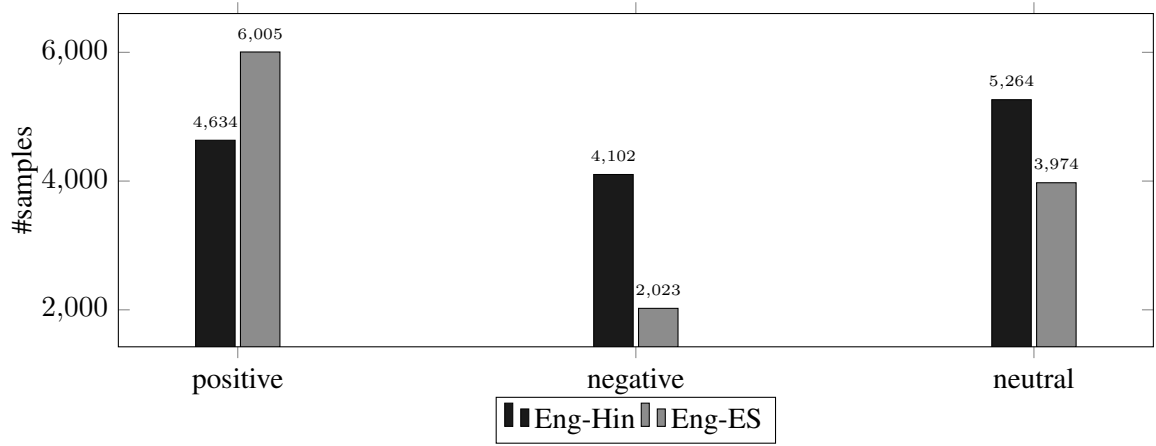
\begin{figure}[!hbtp]
\begin{tikzpicture}
 \begin{axis}[
 width=0.95\textwidth,
 height=6cm,
 ybar,
 enlargelimits=0.15,
 legend style={at={(0.5,-0.15)},
 anchor=north,legend columns=-1},
 ylabel={\#samples},
 symbolic x coords={positive,negative,neutral},
 bar width=3ex,
 xtick=data,
 nodes near coords,
 nodes near coords align={vertical},
 ]
 \addplot[color=black, fill=black!90,font=\tiny] coordinates {(positive,4634) (negative,4102) (neutral,5264)};
 \addplot[color=black, fill=gray!90, font=\tiny] coordinates {(positive,6005) (negative,2023) (neutral,3974)};
 \legend{Eng-Hin,Eng-ES,}
 \end{axis}
\end{tikzpicture}
\caption{Categorical Label Frequency in Hinglish and Spanglish Training Dataset. \label{fig:overflow}}
\end{figure}

The data was provided in the CONLL format with each tweet having a sentiment of ``positive'', ``negative'' or ``neutral''. Figure~\ref{fig:overflow} gives us an overview of the class distribution of these datasets. As we can see the ``Spanglish'' dataset is unbalanced with the majority of the samples being in the ``positive'' and ``neutral'' datasets. From the examples in Table~\ref{table:eg} we can see that the organizers have also provided word level information, with the classes \textit{Eng (English)}, \textit{Es (Spanish)}, \textit{Hin (Hindi)}, \textit{mixed}, and \textit{univ} (\emph{e.g.}, universal symbols, @ mentions, hashtags).

%Apart from sentence level tags the dataset contains word-level as seen in Table~\ref{table:eg} the language tags which are en (English), spa (Spanish), hi (Hindi), mixed, and univ (\emph{e.g.}, symbols, @ mentions, hashtags). 
% \begin{figure}[ht!]
% \centering
% \includegraphics[width=75mm]{chart.png}
% \caption{Categorical Label Frequency in Hinglish dataset \label{overflow}}
% \end{figure}

\section{Methodology}
We extracted different hand engineered features like word $n$-grams, sentiment polarity, punctuation frequency, word frequency, emoji frequency and profanity level from tweets. We weighed $n$-gram features using their term frequency - inverse document frequency (TF-IDF) scores. We then built a Logistic Regression classification model using Scikit-learn~\cite{scikit-learn} library.  %We tuned the we apply a grid search cross-validation over the values {10, 1, 0.1, 1e-02, 1e-03, 1e-04, 1e-05, 1e-06} during the training phase to tune the penalty factor parameter. 
% This section demonstrates how we can extract context rich features from the tweets and apply them to a classifier to differentiate between ``positive", ``negative" and ``neutral" classes. The features we use are: word $n$-grams, sentiment polarity, punctuation frequency, word frequency, emoji frequency and profanity level. With these hand-engineered features we built a Logistic Regression classification model with linear kernel using the implementation of Scikit-learn~\cite{scikit-learn}. We apply the linear kernel as most of
% the text classification problems are linearly separable~\cite{10.1007/BFb0026683}. Finally we apply a grid search cross-validation over the values {10, 1, 0.1, 1e-02, 1e-03, 1e-04, 1e-05, 1e-06} during the training phase to tune the penalty factor parameter. 

\subsection{Pre-processing}

Prior to using the datasets we explored a variety of pre-processing techniques to reduce the noise in the data as they are obtained from online social media and include additional textual features. We used the following pre-processing techniques.                                                            
\begin{itemize}

 \item  \noindent{\bf Abbreviation Expansion}: In many cases users on online social media tend to use informal slang and abbreviations. For instance, we expanded common abbreviations like ``DM'' to ``direct message'' using the Python regular expression library.

 \item  \noindent{\bf Repetition}: As communication on social media websites is informal, there are  cases where letters in words are repeated such as “heelloo”. We normalize these words by using the Python regular expressions library.

%s\noindent{\bf Emoticons }: Emoticons are important to the sentiment analysis task as it can convey the overall emotion of a tweets, for example ":)" can represent happiness. We remove emoticons from the text but keep a record of it and use it as a feature. 

 \item  \noindent{\bf Usernames and Handles}: We use the Python regular expression library to develop a manually defined function to remove user handles for example (@Matt), URLs (\url{https:twitter.com}) and similar links to external images.

\end{itemize}

\subsection{Hand-crafted Features}\label{sec:features}
We explored and used different hand-crafted features to build our machine learning model. Our hand-crafted features are described below:

\begin{itemize}

  \item  \noindent{\bf Lexical Features}: We extracted word $n$-grams ($n=1,2,3$) from the tweets as they are strong lexical representations~\cite{cavnar1994n,McNamee2004,sureka}. A word $n$-gram is simply a sequence of $n$ words, for example ``hello again'' is a bi-gram or $2$-gram. 

  \item  \noindent{\bf Sentiment Lexicon Features}: We extracted the sentiment polarity for English language tokens using Valence Aware Dictionary and sEntiment Reasoner (VADER)~\cite{vader}. VADER is a lexicon and rule-based sentiment analysis tool that is specifically attuned to sentiments expressed in social media. This feature provides strong clues about the sentiment of the tweets by explicitly discriminating ``positive'' or ``negative'' words in the tweet.

  \item  \noindent{\bf Emoji Features}: Emoji are frequently used in online discourse and give us a hint about the tone of the tweets. We used a rule-based dictionary to formalize their polarity as ``happy'', ``sad'', and ``neutral''. We constructed this lexicon using the Emoji Sentiment Ranking~\cite{10.1371/journal.pone.0144296}. 

  \item  \noindent{\bf Profanity Features}: Profane words are correlated with negative connotations. We used the presence or absence of profane words as a feature to discriminate between the three classes. 

  \item  \noindent{\bf Tweet Metadata Features}: Finally we looked into using metadata about the tweet. We extracted repetition, punctuation count, and length and used them as features.

\end{itemize}

\subsection{Experimental Setup}
We used the training and validation splits for Spanglish and Hinglish datasets to build our final classifier. We experimented with standard machine learning classifiers like Logistic Regression, Support Vector Machines, Random Forests with our hand-crafted features defined in Section~\ref{sec:features}. We tuned the $C$ hyperparameter of the Logistic Regression and Support Vector Machine using an extensive grid search over the range of values \{$10e^{-2}, \ldots, 10$\} on the validation dataset. For our final model, we used the best hyperparameter value on the validation dataset. After a number of trials by using a step size of 0.01 we found that $0.9$ was the best value to use as it gave us the best results. Additionally, based on the size and type of the data we selected the ``Liblinear'' solver, which works in a one vs. rest fashion for multi-class tasks.

Moreover, we ran experiments using deep learning models. For deep learning models, we used BERT~\cite{devlin-etal-2019-bert},  ELMo~\cite{peters-etal-2018-deep}, and GloVe~\cite{pennington-etal-2014-glove} embeddings to initialize the embedding layer, Bidirectional Gated Recurrent Units~\cite{cho-etal-2014-learning} for sequence to sequence encoding, and used self-attention~\cite{DBLP:conf:iclr:LinFSYXZB17} to reduce the sequence of encoded vectors to a single vector for representing the tweet. The tweet representations were then used for classification. Additionally, we used the Adam optimizer to train our deep neural network. To select our top candidate we evaluated these models based on the F1 score.

\section{Results and Analysis}

%As outlined in the previous section we base our approach towards exploring feature engineering with linear models and deep learning systems. To select a candidate model we explored a variety of robust classification methods like, Logistic Rregression, Random Forests, Support Vector Machines, Feedforward networks and languages models like BERT. As a common theme we used a 10 fold cross validation along with grid search to tune each model and come up with an average performance. To rank the models we use the average F1 score.

\begin{table}[!htbp]
\centering
\begin{tabular}{lrr}
 \toprule
 Model & Hinglish F1 Score & Spanglish F1 score \\
 \midrule
Logistic Regression & \textbf{0.58} & \textbf{0.55}\\ 
Support Vector Machines & 0.49&0.50\\ 
Feed-forward Networks & 0.56&0.53\\ 
Random Forest & 0.48&0.46\\
BERT Language Model & 0.56&0.48\\
GloVe + ELMo Language Model & 0.56&0.54\\ \hline 
Organizer Baseline & 0.58&0.49\\
 \bottomrule
 \end{tabular}
\caption{Model Results for Hinglish and Spanglish Validation dataset}
\label{table:valresults}

%  \begin{tabular}{lr}

%  \toprule
%  Model & F1 Score \\
%  \midrule
% Logistic Regressionc\\ 
% Support Vector Machines & 0.50\\ 
% Feed-forward Networks & 0.53\\ 
% Random Forest & 0.46\\
% Bert Language Model & 0.48\\
% Glove + Elmo Language Model & 0.54\\
% Organizer Baseline & 0.49\\

%  \bottomrule

%  \end{tabular}

\end{table}

Table~\ref{table:valresults} shows us the results of a variety of classification models applied to the Hinglish and Spanglish validation datasets. As we can see the ``Logistic Regression'' model performed the best giving us a F1 score of 0.58 for the Hinglish validation dataset and 0.55 for the Spanglish validation dataset. It is interesting to see that the use of traditional hand-engineered features performed better than the state of the art deep learning approaches. We suspect that the limited amount of data negatively affects deep learning approaches and makes it conducive to use linear classification models. As a result, we chose to use the ``Logistic Regression'' classifier on the final dataset. 

%Table~\ref{table:valresults} shows us the results from our intial experiments on the validation dataset. As we can see, the Logistic Regression model performed the best giving us a score of 0.58  matching the organizer baseline of 0.58.  Again, Table~\ref{table:valresults} shows us the results of models on the "Spanglish" dataset which again gave us strong results with the Logistic Regression model which had an average score of 0.55 which was higher than the organizer baseline of 0.49.

After the release of the test dataset, we merged the training dataset with the validation dataset and trained our model with the best hyperparameters. For the Hinglish test dataset, we got a significantly higher score of 0.65, which matches the organizer baseline of 0.65. Additionally, for the Spanglish dataset, we got a score of 0.63 which is a lower than the organizer baseline of 0.65.

\subsection{Error Analysis}

\begin{table}[!htbp]
 \centering
 \begin{tabular}{ p{10cm}ll}
 \toprule
 Text & Gold & Prediction \\
 \midrule
 Kameeny loogo ko 
 justice system sy 
 wessy he nikal 
 deena chaheye khud 
 dafa hona bhi 
 theek hy . & Negative & Negative \\
 \hline
 Dear all of every 
 one so I am help
 you me Vishnu 
 Sharma gwalior 
 madhyapradesh se
 hoo My mere 
 teen better he Jo. & Positive & Positive \\
 \hline
`bht moody ho
aajkal . hope all
is well & Negative & Neutral \\
 \hline
`receptionist sit 
wherever you ' d like
me thank you I ' ll 
be in my car' & Neutral & Neutral \\

 \hline
I'm  poor 
poor-ever 
happy kase 
God  is  with \includegraphics[height=.8em]{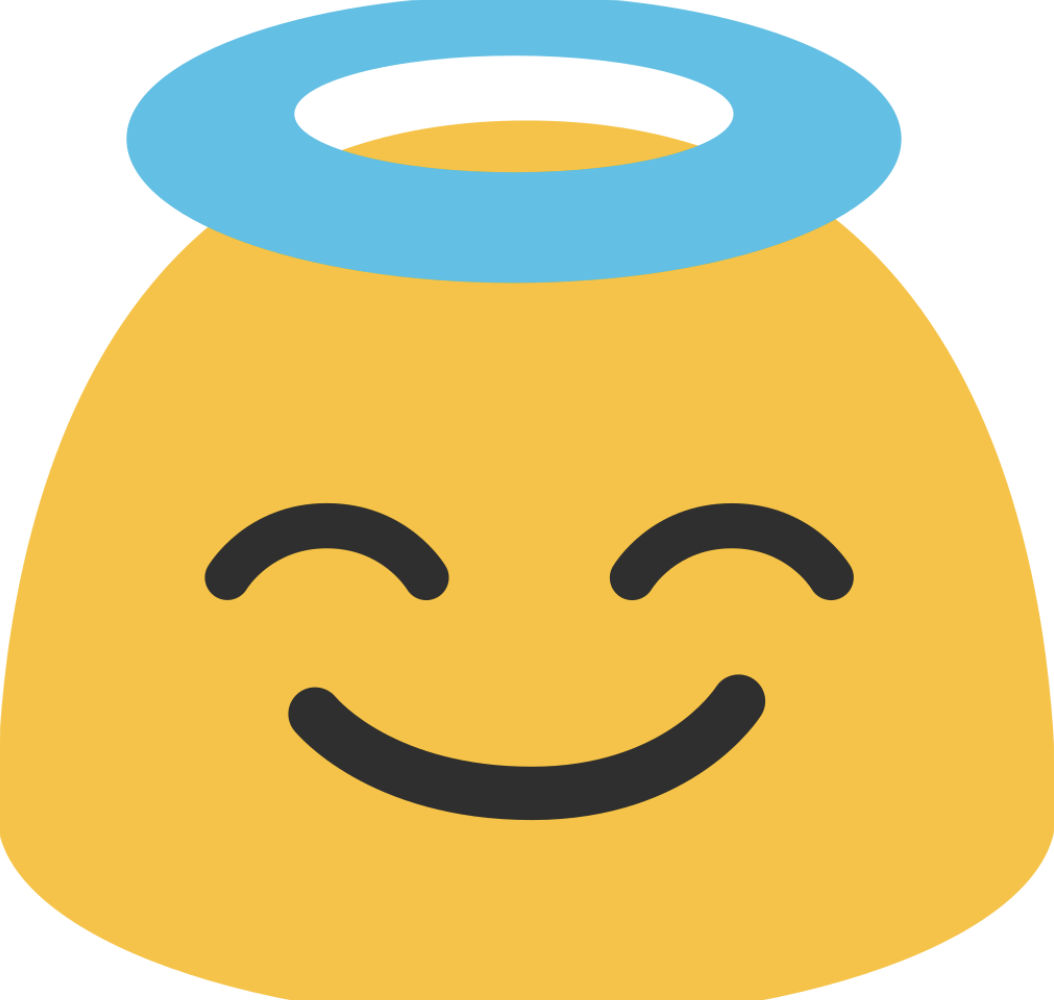}
& Negative & Positive\\

 \bottomrule
 \end{tabular}
 \caption{Sample Predictions from the Hinglish Validation Dataset }
 \label{table:exampleresult}
\end{table}

Table~\ref{table:exampleresult} shows us some of the label predictions from the dataset. In many cases, words associated with profanity and emoticons help the model decide between classes like positive and negative but these do not help the model distinguish between ``neutral'' and ``positive'' or ``negative'' tweets as we do not have a list of words specific to a ``neutral'' class. As we can see in the third row in Table~\ref{table:exampleresult} our model gives us an insight into incorrectly classified results. The model gives us a prediction of ``Neutral'' where the true label is ``Negative''. As seen in the last row of Table~\ref{table:exampleresult} when the tweet is ambiguous the emoji contributes significantly to the prediction. In this case, our prediction is ``Positive'' but the gold label is ``Negative''. We suspect the presence of emoji with positive polarity signaled the model to predict ``Positive'' label. While this is an incorrect prediction we can perhaps infer that the dataset has incorrectly labeled samples.

\section{Conclusions and Future Work}
In this paper, we demonstrated our system for performing sentiment analysis on code-mixed tweets. We used a variety of lexical, metadata, and sentiment-based features. We used a Logistic Regression Classifier with these features to classify the tweets as ``positive'', ``negative'' and ``neutral''. We demonstrate how a lightweight and memory-efficient model with prior extracted features can be competitive with state of the art Language Models. 
In the future, we would look into combining hand-engineered lexical, sentiment, and metadata features with the representations learned from Convolutional Neural Networks (CNN) and Bidirectional Gated Recurrent Unit (Bi-GRU) having attention model applied on top~\cite{kar-etal-2017-ritual}. Recent developments in NLP research have pointed to the combination of deep learning representations as a strong approach to gain better results.

\bibliographystyle{coling}
\bibliography{coling2020}

\end{document}